\documentclass[runningheads]{llncs}
\usepackage[T1]{fontenc}
\usepackage{graphicx}
\usepackage[utf8]{inputenc} 
\usepackage[T1]{fontenc}    
\usepackage{hyperref}       
\usepackage{url}            
\usepackage{booktabs}       
\usepackage{amsfonts}       
\usepackage{nicefrac}       
\usepackage{microtype}      
\usepackage{xcolor}  
\usepackage{tabularx}
\usepackage[utf8]{inputenc}
\usepackage{booktabs}
\usepackage{multirow}
\usepackage{caption}
\usepackage{graphicx}
\usepackage{amsmath}

\usepackage{graphicx}
\usepackage{booktabs}
\usepackage{array}
\usepackage{caption}
\usepackage[table]{xcolor}

\begin{document}
\title{NSMQ Riddles: A Benchmark of Scientific and Mathematical Riddles for Quizzing Large Language Models}
\titlerunning{NSMQ Riddles}
%

\author{George Boateng\inst{1,3} \and
Naafi Dasana Ibrahim \inst{3}\and
Samuel John \inst{3}\and
Philemon Badu \inst{3}\and
Patrick Agyeman-Budu \inst{3}\and
Jonathan Abrefah Mensah\inst{3}\and
Kevin Takyi Yeboah \inst{3,4}\and
William Edor \inst{3}\and
Andrew Kojo Mensah-Onumah \inst{3}\and
Nana Sam Yeboah \inst{3}\and \\
Victor Kumbol \inst{2,3}}





\authorrunning{G. Boateng et al.}

\institute{ETH Zurich, Switzerland\\
\and
Charité - Universitätsmedizin Berlin, Germany\\
\and
Kwame AI Inc., U.S.\\
\and
Ashesi University, Ghana\\
}


\maketitle 


\begin{abstract}
Large  Language Models (LLMs) have shown good performance on various science educational benchmarks, demonstrating their potential for use in science and mathematics education. Yet, LLMs tend to be evaluated on science and mathematical educational datasets from the Western world, with an underrepresentation of datasets from the Global South. Furthermore, they tend to have multiple-choice answer options that are trivial to evaluate. In this work, we present NSMQ Riddles, a novel benchmark of Scientific and Mathematical Riddles from Ghana’s National Science and Maths Quiz (NSMQ) competition to evaluate LLMs. The NSMQ is an annual live TV competition for senior secondary school students in Ghana that brings together the smartest high school students in Ghana who compete in teams of 2 by answering questions in biology, chemistry, physics, and math over five rounds and five stages until a winning team is crowned for that year. NSMQ Riddles consists of 11 years of riddle questions (n=1.8K) from the 5th round, with each riddle containing a minimum of 3 clues. Students compete to be the first to guess the answer on any of the clues, with earlier clues being vague and also fetching more points. The answers are usually a number, word, or short phrase, allowing for automatic evaluation.  We evaluated state-of-the-art models: closed (GPT-5.4, Gemini 3.1 Pro, Claude Opus 4.6) and open models (Kimi-K2.5, DeepSeek-V3.1, GPT-OSS-120B) with high and low reasoning settings. Our evaluation shows that the dataset is challenging even for state-of-the-art LLMs, which performed worse than the best student contestants. This work contributes a novel and challenging benchmark for scientific and mathematical reasoning from the Global South towards enabling a true global benchmarking of LLMs’ capabilities for science and mathematics education.

\keywords{Generative AI  \and LLMs \and Question Answering \and STEM \and Science \and Mathematics \and Riddles}
\end{abstract}

\section{Introduction}
Large Language Models (LLMs) have demonstrated good performance on various science and mathematics educational benchmarks \cite{hendrycks2020,cobbe2021}, showcasing their potential in science and mathematics education. These models excel at tasks ranging from answering questions to generating educational content. Nevertheless, a significant gap exists in the evaluation of these models using educational datasets from the Global South, with limited datasets such as \cite{zhang2024}. Predominantly, LLMs are assessed with educational datasets from Western contexts \cite{mihaylov2018,Clark2018,Kembhavi2017,Lu2022}, failing to adequately represent the diverse educational challenges found in other parts of the world, particularly in Africa. For instance, GPT-4’s release in March 2023 included various academic benchmarks, yet none were from Africa \cite{achiam2023}. This oversight highlights a broader issue: the marginalization of African educational contexts in evaluating the capabilities of LLMs. This geographic bias in dataset selection limits the understanding of LLMs' true capabilities and their potential global impact. This underrepresentation not only perpetuates the marginalization of African educational contexts in AI research but also hinders the development of AI solutions tailored to the unique needs of African students. By focusing predominantly on Western datasets, current benchmarks fail to capture the full spectrum of educational scenarios, thus limiting the applicability and fairness of AI advancements in education globally. Addressing this problem is crucial for several reasons. Firstly, ensuring that AI technologies are inclusive and equitable is essential for providing educational benefits across diverse contexts. Secondly, incorporating datasets from the Global South enables researchers to develop more robust and generalizable AI models that perform well across various cultural and educational settings. Lastly, addressing this gap can help reduce educational inequities by leveraging AI to support and enhance learning in underrepresented regions.

In this work, we introduce NSMQ Riddles, a novel benchmark comprising Scientific and Mathematical Riddles from Ghana’s National Science and Maths Quiz (NSMQ) competition, aimed at evaluating LLMs in an educational context. The NSMQ is an annual live TV competition in Ghana that brings together the smartest high school students in Ghana who compete in teams of 2 by answering questions in biology, chemistry, physics, and math over five rounds and five stages until a winning team is crowned for that year.  \cite{nsmq,boateng2023nsmq}. The questions for the NSMQ are set by academics at the University of Ghana and and covers high-school level biology, chemistry, physics, and mathematics questions aligned with the West African Senior School Certificate Examination (WASSCE) curriculum which is used by Ghana, Nigeria, Gambia, Sierra Leone, Liberia and The Gambia \cite{boateng2023nsmq}.  This work focuses on round 5 — Riddles, and it is a part of the Brilla AI \footnote{\url{https://brilla-ai.org/}} project that is building an AI contestant to win the NSMQ \cite{boateng2024}.  This round, the final one, is arguably the most exciting as the competition's winner is generally determined by the performance in the round. In the Riddles round, students answer riddles across Biology, Chemistry, Physics, and Mathematics. Three (3) or more clues are read to the teams that compete against each other to be first to provide an answer (usually a word or a phrase) by ringing their bell. The clues start vague and get more specific, which make earlier clues more challenging. To make it more exciting and encourage educated risk-taking, answering on the 1st clue fetches 5 points, on the 2nd clue — 4 points, and on the 3rd or any clue thereafter, 3 points. There are 4 riddles for each contest with each riddle focusing on one of the 4 subjects. Speed and accuracy are key to winning the Riddles round. An example riddle with clues and the answer is as follows (see a live example here \footnote{Video Example of Riddle:  \url{https://www.youtube.com/watch?v=pE42doghgXw}}). 

\textbf{Question}
\begin{enumerate}
    \item I am a named principle in physics. 
    \item I am a definite principle, my name notwithstanding.
    \item I have been interpreted to mean a natural restriction on the pairs of measurements that can be simultaneously performed.
    \item However, I only give bounds.
    \item Position and momentum are the common candidates for a discussion of me	
    \item but I have more general applicability when I am formulated appropriately. Who am I? 
\end{enumerate}
\textbf{Answer:} Heisenberg Uncertainty Principle. 

NSMQ Riddles includes 11 years of riddle questions (n=1.8K) from the fifth round, each containing a minimum of three clues. Answers are typically a number, word, or short phrase, allowing for automatic evaluation. We evaluated current state-of-the-art models: closed (GPT-5.4, Gemini 3.1 Pro, Claude Opus 4.6) and open models (Kimi-K2.5, DeepSeek-V3.1, GPT-OSS-120B) with high and low reasoning settings. We compared their performance to the retrospective performance of the best student contestants in the competition for those riddles. Our results indicate that the dataset poses challenges even for these advanced models, which performed worse than the best student contestants.

Our contributions are (1) a novel dataset of 1.8K scientific and mathematical riddles from the Global South and (2) an evaluation of closed and open state-of-the-art LLMs on a subset of the dataset. To the best of our knowledge, this is the first scientific and mathematical riddles dataset. This work contributes a novel benchmark for scientific and mathematical reasoning from the Global South, promoting a truly global evaluation of LLMs’ capabilities in education.

\section{Related Work}
We describe related work on (1) riddles datasets and (2) science and mathematics question-answering datasets, along with how our work differs.

\subsection{Riddles Datasets}
The creation of datasets focused on riddles that require nontrivial commonsense and counterfactual reasoning has gained traction in recent years, driven by the need to evaluate the capabilities of natural language processing models beyond factual question-answering tasks. Several efforts have been undertaken, providing valuable insights and benchmarks for this work.  One notable dataset is RiddleSense, a collection of human-written riddles designed to assess the commonsense reasoning abilities of language models with multiple-choice answer options obtained by crowdsourcing distractors \cite{Lin2021}. Another dataset is the BiRdQA dataset, which offers bilingual riddles in English and Chinese with multiple-choice answer options (distractors generated automatically) accompanied by Wikipedia introductions for each candidate answer \cite{Zhang2022}. This dataset has facilitated cross-lingual studies, enabling comparative analyses across different languages and cultural contexts. NSMQ Riddles differs from these riddles datasets in its (1) focus on science and mathematics educational question answering, (2) the complexity of each riddle containing a minimum of 3 clues and an average of 5 clues, and (3) the use of short answer responses rather than multiple-choice answer options.

\subsection{Science and Mathematics Datasets}
Several datasets exist for science educational question answering such as OpenBookQA \cite{mihaylov2018}, ARC \cite{Clark2018}, TQA \cite{Kembhavi2017}, ScienceQA \cite{Lu2022}, SciBench \cite{wang2024}, MMLU-STEM \cite{hendrycks2020}, GPQA \cite{rein2024}(tasks focused on science and engineering subjects) and mathematics question answering such as MATH \cite{hendrycks2021}, GSM8K \cite{cobbe2021}, MathArena\cite{balunovic2025}. These datasets span various educational levels from elementary school through to college level, with varying levels of difficulty. Most of these datasets use multiple choice answer format, which allows for simple evaluation. NSMQ Riddles differs from these datasets primarily by the riddle structure of questions, which require complex reasoning across clues to arrive at an answer, goes beyond multiple-choice answer options through its use of short answer responses, and, importantly, comes from an educational context in the Global South (West Africa), which tends to be underrepresented in these benchmarks. 


\section{NSMQ Riddles}
\subsection{Overview}
The NSMQ Riddles dataset \footnote{For access to the dataset for benchmarking, send an email to nsmq.kwame.ai@gmail.com} consists of 1,840 riddle questions comprising 11 years of questions from round 5 of the NSMQ, ranging from 2009 to 2021 (excluding 2010 and 2011 when the contests did not occur). In general, the questions get more difficult in later stages of the competition, with the grand finale being the most challenging. The dataset features an almost equal distribution across biology (25.7\%), chemistry (28.8\%), physics (26.4\%), and mathematics (24.5\%); note that each riddle could have multiple subject tags and so the total percentage is over 100\%. In the Riddles round, each question is made up of three (3) or more clues read aloud to competing teams. The teams race to be the first to ring their bell and give the correct answer, usually a word or phrase. Clues begin broadly and become progressively more specific. To add excitement and reward smart risk-taking, a correct answer on the first clue earns 5 points, the second clue earns 4 points, and the third or any subsequent clue earns 3 points. Each contest features four riddles, one from each of the four subjects. Speed and accuracy are essential to winning the Riddles round. Several examples with one per subject are shown in Table \ref{tab:example_riddles}. Here is an example riddle with clues and the corresponding answer.

\textbf{Question}
\begin{enumerate}
    \item You have to look down below the horizontal to view me.
    \item I am always downcast and feeling depressed.
    \item I am an angle.
    \item To be precise, I am a type of angle. 
    \item I am the angle between the horizontal and a line from an observer to an object below the eye level of the observer. Who am I?    
\end{enumerate}
\textbf{Answer:} Angle of depression. 

Furthermore, the dataset contains a metadata sheet with information about the contests for 2019 (156 riddles) and 2020 (160 riddles), such as the date of the contest, competing schools, and the school that answered each riddle and at which clue, which we used to evaluate the performance of the best contesting student teams for each riddle. Note that in the metadata sheet, 4 riddles in 2019 have incomplete data due to glitches in the video recordings. 


\begin{table}[h]
\caption{Examples of Riddles across Physics, Mathematics, Chemistry and Biology}
\label{tab:example_riddles}
\begin{tabularx}{\textwidth}{|X|l|l|}
\hline
Riddle & Answer & Subject \\
\hline
I am a property of an object. I am a mechanical property. In SI base units, I am measured in kgm2. For a collection of particles, I am the sum of the products of the mass of each particle and its distance squared from an axis. I am a second moment of mass. & Moment of inertia & Physics \\
\hline
I am a quadrilateral. My diagonals bisect the vertex angles. My diagonals are perpendicular. My diagonals bisect each other. My diagonals are congruent.  & Square & Mathematics \\
\hline
I am relatively small and cute. My central atom has sp3-hybridised orbitals. My shape is inevitably tetrahedral. I am a greenhouse gas whether I am in my pure state or oxidised. With my penta-atomic molecules, I am the smallest member of a homologous series. & Methane & Chemistry \\
\hline
I am a small hole behind each eye that opens to the mouth in some fishes. In tadpoles I am the excurrent opening from the gill chamber. In elasmobranch and ganoid fishes I am derived from the gills, and am used as a water passageway during respiration. I am usually found on certain thoracic and abdominal segments. Insects and some more derived spiders have me on their exoskeletons to allow air to enter the trachea. I am a respiratory opening found on the thorax and abdomen of insects. & Spiracle & Biology \\
\hline
\end{tabularx}
\end{table}

\subsection{Data Curation and Preprocessing}
We obtained publicly available video recordings of the competition on YouTube and the digital version of the questions and answers (which contained metadata such as subjects of each question, contest number, and year) through our collaboration with a high school that participates in the competition. We parsed them and reformatted them into CSVs for usage. We obtained the questions in PDF format and used Mathpix \cite{mathpix}, a commercial Optical Character Recognition tool, to convert them into Markdown-formatted text files that transformed equations and formulas into LaTeX. Since OCR tools like Mathpix can occasionally introduce transcription errors, we manually reviewed each converted document by comparing the OCR output against the original PDF and correcting any discrepancies. We also used StackEdit to clean up any formatting issues left over from the conversion process. To enable automatic evaluation, we manually created variants of the ground truth answer(s) that included stripping off Markdown, separating correct answer options (e.g., Sodium palmitate OR Sodium hexadecanoate), and providing alternative forms of answers such as writing out the full chemical name if the symbolic form is the ground truth answer, and vice versa (e.g., hydrogen and h2).  For example, the following ground truth answer originally in LaTeX, $SO_{2}$, results in the following alternative answers: SO2 and Sulphur (IV) Oxide. We then wrote code that reformatted the questions and answers text file into a CSV. The CSV had columns “Clue 1” to “Clue 9” for all clues, an “Answer” column for the original, ground truth answer(s), columns “Answer with Markdown 1”, and  “Answer with Markdown 2” which contained all the different versions of the answer(s) in LaTeX and “Answer 1”, “Answer 2”, “Answer 3” and “Answer 4” which contained different versions of the ground truth answer(s) that have their LaTeX stripped off or answers that are in plain text form. Additional columns were added to track information like subject, contest number, and year of the contest. We applied the following preprocessing steps and created riddle-answer mappings: converted all clue texts and answers to lowercase, removed punctuation, fixed whitespace, and removed articles (e.g., “the”, “a”, “an”). 

To create the metadata sheet, we searched for images or PDFs about the contests on social media and blogs, as well as the Facebook and YouTube links to the previously livestreamed contests. We then created a Google sheet and populated it with information about each contest, such as the contest date and the names of the competing schools. We then watched each contest to provide information about the marks they acquired at the end of each contest and for the riddles round in particular, information such as which school answered each riddle and at which clue, which was used to calculate the performance of the best human contestants for each riddle. Resource constraints allowed for the complete curation of metadata for only the years 2019 and 2020.

\section{Evaluation}
We evaluated closed and open state-of-the-art models using an evaluation setup similar to \cite{boateng2024}. We used a well-developed prompt to generate an answer given a set of riddle clues. We used prompt engineering to develop a high-quality prompt that takes the clues as part of the prompt. The prompt (1) asked the model to take on the role of an expert — a science prodigy, (2) used Chain-of-Thought (CoT) prompting by asking it to reason through the clues, (3) stated a penalty for deviating from the instruction to provide a short answer, (4) used few-shots learning by giving an example of a riddle and an answer, and (5) asked for a structured output — JSON.

We used Exact Match (EM) accuracy as the primary metric and a secondary metric, Fuzzy Match (FM) accuracy. With EM, we search for instances where the generated answer exactly matches the ground truth answer for a given riddle or the alternative answers if there are any. With FM, we perform a more relaxed comparison and check if any of the ground truth answers is a substring of the answer generated by the model and vice versa. For example, (ground truth: tissue, model answer: tissues, will yield True for fuzzy match). We opt for EM as our primary evaluation metric, given the quiz’s nature, which requires that the answers provided by the competing teams precisely match the term the riddle points to. Additionally, we employ FM to gauge the generated answers' proximity to the ground truth.

For all evaluations of the open models, we used the Together AI API, and for the closed models, we used the LangChain \cite{langchain} wrapper for the OpenAI API. We set the temperature to 1 in all instances and max\_tokens = 20000 for models that require the parameter. We performed 2 sets of evaluations: Offline, which assumes the model has access to all the clues for each riddle and generates an answer, and Real-Time Proxy, which approximates the real-time context of the quiz by passing chunks of riddle clues as input to the model to decide whether or not to answer. We evaluated the models’ performance on riddles from only 2019, given that we had limited compute resources and annotations to perform the comparison with students.

\subsection{Offline Evaluation}
For the offline evaluation, we provide each model with comprehensive access to all the clues necessary to answer the riddle. Specifically, we concatenate all the clues of a riddle into one question string, format this into our prompt template (Table \ref{tab:offline_evaluation_prompt}), and then feed this prompt to the model. This approach ensures the model has all available information for a given riddle, allowing it to reason through the clues thoroughly. The model then analyzes the provided information and generates an answer to the riddle. In addition to evaluating the overall performance, we show a detailed breakdown of results by subject. This breakdown allows us to assess how well the model performs in different areas of knowledge, providing insights into its strengths and weaknesses across various disciplines.

\begin{table}[]
\centering
\caption{Prompt for Offline Evaluation}
\label{tab:offline_evaluation_prompt}
\begin{tabular}{|p{0.2\textwidth}|p{0.8\textwidth}|}
\hline
System Message &
  \begin{tabular}[c]{@{}p{0.8\textwidth}@{}}You are a science prodigy currently competing in a National Science competition.\\ You are now in the fifth round, where you must provide an answer to a riddle.\\ Remember, your answer should consist of just the term the riddle is pointing to, and nothing else.\\ Adding additional text will result in point deductions.\\ \\ Your response should be a single serializable JSON Object with the following properties:\\ - answer, str: Your answer to the riddle.\\ \\ Do not add any additional text, except the JSON in the format specified.\\ \\ Here's an example to guide you:\\ Riddle: You might think I am a rather unstable character because I never stay at one place.\\ However, my motion obeys strict rules and I always return to where I started.\\ And even if I have to leave that spot again I do it in strict accordance to time.\\ I can be named in electrical and mechanical contexts.\\ In all cases, I obey the same mathematical rules.\\ In order to fully analyze me you would think about a stiffness or force constant restoring force and angular frequency.\\ \\ Your response:\\ \{'answer': 'oscillator'\}\end{tabular} \\ \hline
Human Message &
  \begin{tabular}[c]{@{}p{0.8\textwidth}@{}}You are presented with the riddle below.\\ Analyze it carefully and thoroughly before proceeding to provide your answer.\\ Your response should be in the format specified.\\ The riddle is delimited by triple backticks.\\ \\ Riddle: ```\{riddle\_content\}```\end{tabular} \\ \hline
\end{tabular}%
\end{table}

\subsection{Real-Time Proxy Evaluation}
In the Real-Time Proxy Evaluation, our objective is to assess the models' performance as if the models were competing in a live NSMQ competition. In the live deployment of an AI contestant during the 2023 NSMQ finals, the authors transmitted five-second audio transcripts to the AI system for analysis and response \cite{boateng2023nsmqai,boateng2024}. We determined that each five-second chunk contains an average of 7 words. Consequently, we provide the models with substrings of the riddle consisting of 7 words each. The models were instructed (Table \ref{tab:real-time_evaluation_prompt}) to use the information they have at hand to produce an answer and a confidence of ‘sure’ or ‘not sure’ to indicate their confidence in their answer. If confident, we take the generated answer as the model’s final response to the riddle. If not, we transmit the subsequent substring of 7 words, concatenate it with the previously received information, and prompt the model to generate an answer with the updated clues. This process continues until the model is confident in its answer or it reaches the final chunk of the riddle. In the latter case, we take the model's predicted answer as its final response regardless of its confidence in the answer. We then calculated EM and FM accuracy. We also computed the points that would have been obtained by each model based on the clue it answered on. We also calculated points for students using the sheet that contains detailed information on which clue the students answered on for each riddle. Note that this approach of calculating points is a proxy, given that the actual points obtained by the LLM or student for each riddle will be affected by whether the students or the LLM answer first in a real competition. That evaluation will be done in future work when we obtain the exact timestamps for when students answered and can perform a real-time evaluation. We compared the LLM results with the retrospective real-world performance of the best-performing student teams for the riddles in those years’ competition. We also computed the average number of clues it takes for each model to answer, and compared it to the students.

\begin{table}[]
\centering
\caption{Prompt For Real-Time Proxy Evaluation}
\label{tab:real-time_evaluation_prompt}
\begin{tabular}{|p{0.2\textwidth}|p{0.8\textwidth}|}
\hline
System Message & \begin{tabular}[c]{@{}p{0.8\textwidth}@{}}You are a science prodigy currently competing in a National Science competition.\\ You are now in the fifth round, where you must provide an answer to a riddle.\\ You are given a set of clues at a time, and at each point, you must determine what you think the answer is,\\ and also if you are sure of the answer given the information you have at hand.\\ \\ You response should be a single serialiazable JSON Object with the following properties:\\ - answer, str: Your answer to the riddle.\\ - confidence, str: 'sure' if you are sure of your answer and 'not sure' if you are unsure of the answer.\\ \\ Here's an example to guide you:\\ Riddle: You might think I am a rather unstable character because I never stay at one place.\\ However, my motion obeys strict rules and I always return to where I started.\\ And even if i have to leave that spot again I do it in strict accordance to time.\\ I can be named in electrical and mechanical contexts\\ In all cases, I obey the same mathematical rules.\\ In order to fully analyse me you would think about a stiffness or force constant restoring force and angular frequency.\\ \\ Your response:\\ \{"answer": "oscillator", "confidence": "sure"\}\end{tabular} \\ \hline
Human Message  & \begin{tabular}[c]{@{}p{0.8\textwidth}@{}}You are presented with the riddle below.\\ Analyze it carefully and thoroughly before proceeding to provide you answer.\\ Your response should be in the format specified.\\ The riddle is delimited by triple backticks.\\ \\ Riddle: ```\{riddle\_content\}```\end{tabular}\\ \hline
\end{tabular}%
\end{table}

\section{Results and Discussion}
We present and discuss the results of the offline and real-time proxy evaluation.

\subsection{Offline Evaluation Results}
The results for the Offline Evaluation are shown in Tables \ref{tab:offline_all_clues_combined}  and \ref{tab:offline_all_clues_subject_breakdown}, with the best results in each column in bold for closed and open models. Our evaluation showed that closed models generally performed better than open models, with GPT-5.4 (with reasoning=high) performing the best with 86.54\% on EM accuracy. Furthermore, higher reasoning settings resulted in better performance, indicating that more reasoning benefits scientific and mathematical question answering. As expected, FM accuracy results are higher than EM, indicating that the model’s answers were not far off from the expected answers, pointing to room for improvement. 

Across all models, our evaluation indicated that they performed quite well across all subjects except Chemistry, which posed a significant challenge, with the best models performing equal to or below 75\% EM accuracy compared to the other subjects, which had the best models above 90\%. A significant challenge for the LLMs on the subject, as mentioned earlier, was their handling of organic compounds and complex terminology. The models generally oversimplified their answers. For example, the models would answer "alkane" instead of the specific compound "Cyclohexane". The models also frequently conflated related concepts, like answering "kinetic theory of gases" instead of the specific "Charles' Law", or naming a reaction mechanism like "free radical substitution" instead of the specific reaction "Chlorination of methane". For organic compounds and substances, they sometimes provided answers from the same general family or class rather than the specific item, such as answering "amino acid" instead of "2-Aminoethanoic acid" or "base" instead of "Milk of Magnesia". This demonstrates a lack of nuanced understanding of the core concepts, suggesting the need for further improvement in this domain.

\begin{table}[h!]
\centering
\caption{Offline Evaluation Results Using All Clues for 2019 NSMQ Riddles (Number of Riddles = 156)}
\label{tab:offline_all_clues_combined}
\resizebox{0.8\textwidth}{!}{%
\begin{tabular}{|l|c|c|}
\hline
\textbf{Models} & \textbf{EM (\%)} & \textbf{FM (\%)} \\ \hline
\multicolumn{3}{|l|}{\textbf{Closed}} \\ \hline
GPT-5.4 (reasoning effort=low)                        & 85.90 & 90.38 \\ \hline
GPT-5.4 (reasoning effort=high)                       & \textbf{86.54} & \textbf{91.03} \\ \hline
Gemini-3.1 (reasoning effort=low)          & 83.33 & 89.74 \\ \hline
Gemini-3.1 (reasoning effort=high)         & 83.97 & 89.74 \\ \hline
Claude Opus 4.6 (reasoning effort=low)                 & 82.05 & 88.46 \\ \hline
Claude Opus 4.6 (reasoning effort=high)                & 84.62 & 89.74 \\ \hline
\multicolumn{3}{|l|}{\textbf{Open}} \\ \hline
Kimi-K2.5 (reasoning effort=low)                      & 80.77 & \textbf{89.10} \\ \hline
Kimi-K2.5 (reasoning effort=high)                     & 80.77 & 88.46 \\ \hline
DeepSeek-V3.1 (reasoning effort=low)                  & 76.28 & 83.97 \\ \hline
DeepSeek-V3.1 (reasoning effort=high)                 & 76.28 & 83.33 \\ \hline
GPT-OSS-120B (reasoning effort=low)                   & 80.77 & 85.90 \\ \hline
GPT-OSS-120B (reasoning effort=high)                  & \textbf{83.33} & 87.82 \\ \hline
\end{tabular}
}
\end{table}

\begin{table}[h!]
\centering
\caption{Subject Breakdown of Offline Evaluation Results Using All Clues for 2019 NSMQ Riddles (Number of Riddles = 156)}
\label{tab:offline_all_clues_subject_breakdown}
\resizebox{\textwidth}{!}{%
\begin{tabular}{|l|c|c|c|c|c|c|c|c|}
\hline
\textbf{Models} & \multicolumn{4}{c|}{\textbf{EM (\%)}} & \multicolumn{4}{c|}{\textbf{FM (\%)}} \\ \cline{2-9}
 & \textbf{Biology} & \textbf{Chemistry} & \textbf{Physics} & \textbf{Math} & \textbf{Biology} & \textbf{Chemistry} & \textbf{Physics} & \textbf{Math} \\ \hline
\multicolumn{9}{|l|}{\textbf{Closed}} \\ \hline
GPT-5.4 (reasoning effort=low)                  & 89.47 & 72.73 & \textbf{92.11} & \textbf{91.67} & 92.11 & 79.55 & 97.37 & 94.44 \\ \hline
GPT-5.4 (reasoning effort=high)                 & \textbf{92.11} & \textbf{75.00} & 89.47 & \textbf{91.67} & \textbf{94.74} & \textbf{81.82} & 94.74 & \textbf{94.44} \\ \hline
Gemini-3.1 (reasoning effort=low)    & \textbf{92.11} & 65.91 & 86.84 & \textbf{91.67} & \textbf{94.74} & 77.27 & 94.74 & \textbf{94.44} \\ \hline
Gemini-3.1 (reasoning effort=high)   & \textbf{92.11} & 68.18 & 86.84 & \textbf{91.67} & \textbf{94.74} & 77.27 & 94.74 & \textbf{94.44} \\ \hline
Claude Opus 4.6 (reasoning effort=low)           & 86.84 & 63.64 & \textbf{92.11} & 88.89 & \textbf{94.74} & 75.00 & \textbf{97.37} & 91.67 \\ \hline
Claude Opus 4.6 (reasoning effort=high)          & \textbf{92.11} & 72.73 & \textbf{92.11} & 83.33 & \textbf{94.74} & 79.55 & \textbf{97.37} & 91.67 \\ \hline
\multicolumn{9}{|l|}{\textbf{Open}} \\ \hline
Kimi-K2.5 (reasoning effort=low)                & 86.84 & 65.91 & \textbf{86.84} & \textbf{88.89} & \textbf{94.74} & \textbf{77.27} & \textbf{94.74} & \textbf{91.67} \\ \hline
Kimi-K2.5 (reasoning effort=high)               & 86.84 & \textbf{65.91} & \textbf{86.84} & 86.11 & 92.11 & \textbf{77.27} & \textbf{94.74} & \textbf{91.67} \\ \hline
DeepSeek-V3.1 (reasoning effort=low)            & 84.21 & 56.82 & \textbf{86.84} & 80.56 & 89.47 & 70.45 & \textbf{94.74} & 83.33 \\ \hline
DeepSeek-V3.1 (reasoning effort=high)           & 86.84 & 54.55 & \textbf{86.84} & 80.56 & 89.47 & 68.18 & \textbf{94.74} & 83.33 \\ \hline
GPT-OSS-120B (reasoning effort=low)             & 89.47 & 65.91 & \textbf{86.84} & 83.33 & \textbf{94.74} & 68.18 & \textbf{94.74} & 88.89 \\ \hline
GPT-OSS-120B (reasoning effort=high)            & \textbf{92.11} & 68.18 & \textbf{86.84} & \textbf{88.89} & \textbf{94.74} & 75.00 & 92.11 & \textbf{91.67} \\ \hline
\end{tabular}
}
\end{table}

\subsection{Real-Time Proxy Evaluation Results}
The results for the Real-Time Proxy Evaluation are shown in Table \ref{tab:real_time_proxy} with the best results in each column in bold for closed and open models. Similar to the offline evaluation, closed models outperform open models, and higher reasoning efforts outperform lower ones. Noticeably, the performance for offline evaluation drops across models for the real-time proxy evaluation, demonstrating that real-time settings, during which not all the clues will be available, will pose a challenge. GPT-5.4 (reasoning effort=high) outperforms all the other models on both EM and FM accuracy and total points. 

The best performance of LLMs of 75.64\% EM accuracy and 366 points is lower than the best students’ performance of 78.21\% EM accuracy and 377 points. The nature of the quiz does not allow for computing FM for students, as only the exact answers are accepted as correct. We can conclude from the results that at present, the best LLMs perform below the level of the smartest high school students in the NSMQ and will likely lose in the riddle round of the real-world competition against the best students. It is important to note that the LLMs’ performance in the real-world context will be affected by various factors such as audio chunk length, transcription errors, and latency, which may result in worse performance. 

Among the closed models, GPT-5.4 particularly tends to wait the longest before answering, while Claude Opus 4.6 answers the earliest, especially with reasoning effort set to high. This tendency to answer early correlates with Claude Opus 4.6’s worse performance in the real-time proxy test compared to the other models, despite its strong performance in the offline evaluation. For the open-source models, however, Kimi-K2.5 and DeepSeek-V3.1 typically answer only after seeing at least 4 clues on average. GPT-OSS-120B, however, behaves similarly to Claude Opus 4.6, and answers much earlier. Consequently, its performance on the real-time proxy is significantly worse as compared to the others, despite its strong performance in the offline evaluation.

These findings point to the need for efforts to improve LLMs on this type of scientific and mathematical reasoning in the context of uncertainty. Such capabilities could enable the use of LLMs to power AI tutors to better support students' science and maths learning.

\begin{table}[h!]
\centering
\caption{Real-Time Proxy Evaluation Results for 2019 NSMQ Riddles -- Overall Performance and Total Points (Number of Riddles = 156)}
\label{tab:real_time_proxy}
\resizebox{\textwidth}{!}{%
\begin{tabular}{|l|c|c|>{\centering\arraybackslash}p{1.5cm}|>{\centering\arraybackslash}p{1.5cm}|}
\hline
\textbf{Models} & \textbf{EM (\%)} & \textbf{FM (\%)} & \textbf{Total Points} & \textbf{Avg Clue Correct} \\ \hline
\multicolumn{5}{|l|}{\textbf{Closed}} \\ \hline
GPT-5.4 (reasoning effort=low)                    & 73.72 & 79.49 & 359 & 4.426 \\ \hline
GPT-5.4 (reasoning effort=high)                   & \textbf{75.64} & \textbf{81.41} & \textbf{366} & 4.381 \\ \hline
Gemini-3.1 (reasoning effort=low)                 & 71.15 & 78.21 & 349 & 4.144 \\ \hline
Gemini-3.1 (reasoning effort=high)                & 73.72 & 80.13 & 358 & 4.243 \\ \hline
Claude-Opus-4-6 (reasoning effort=low)            & 52.56 & 59.62 & 277 & 3.415 \\ \hline
Claude-Opus-4-6 (reasoning effort=high)           & 51.28 & 60.90 & 273 & 3.188 \\ \hline
\multicolumn{5}{|l|}{\textbf{Open}} \\ \hline
Kimi-K2.5 (reasoning effort=low)                  & 64.10 & 69.23 & 313 & 4.240 \\ \hline
Kimi-K2.5 (reasoning effort=high)                 & \textbf{66.03} & \textbf{73.08} & \textbf{315} & 4.417 \\ \hline
DeepSeek-V3.1 (reasoning effort=low)              & 50.00 & 53.85 & 251 & 4.333 \\ \hline
DeepSeek-V3.1 (reasoning effort=high)             & 51.28 & 56.41 & 252 & 4.425 \\ \hline
GPT-OSS-120B (reasoning effort=low)               & 46.15 & 51.28 & 240 & 3.403 \\ \hline
GPT-OSS-120B (reasoning effort=high)              & 46.79 & 52.56 & 241 & 3.164 \\ \hline
\noalign{\vskip 6pt}
\hline
\rowcolor{gray!20}
\textsc{Students (Human Baseline)} & \textbf{78.21} & \text{N/A} & \textbf{377} & 4.426 \\ \hline
\end{tabular}%
}
\end{table}

\section{Limitations and Future Work}
Our work has a number of limitations. First, our real-time proxy evaluation used only one year of the NSMQ (2019) since we had limited annotations of the metadata needed for the evaluation, as well as limited compute resources. Second, our real-time evaluation was only a proxy, given that we did not have the annotated audio versions of the competition. Third, since the questions appear publicly on YouTube, there is a possibility of data contamination but we did not assess that in this work.

Future work will (1) annotate and evaluate on more years of the riddles dataset, (2) curate, annotate, and perform real-time evaluation using the audio form of the riddles, (3) curate and annotate other rounds of the NSMQ to provide a more challenging and comprehensive benchmark, and (4) perform de-contamination analysis.

\section{Conclusion} 
We presented NSMQ Riddles, a novel benchmark of 1.8K Scientific and Mathematical Riddles from the NSMQ competition for evaluating LLMs in an educational context. We evaluated state-of-the-art models: closed (GPT-5.4, Gemini 3.1 Pro, Claude Opus 4.6) and open models (Kimi-K2.5, DeepSeek-V3.1, GPT-OSS-120B) with high and low reasoning settings using EM and FM accuracy. We evaluated in (1) offline mode (having access to all clues and always attempting to answer) and (2) real-time proxy, where chunks of the clues are given to the model to decide if it should answer with a comparison to the real-world performance of students for those riddles. Overall, closed models and high reasoning modes performed better than open models and lower reasoning modes. The best performance for offline was EM accuracy of 86.54\% (GPT-5.4 with reasoning=high), and for real-time proxy,  EM of 75.64\% and 366 points (GPT-5.4 with reasoning=high). Our evaluation shows that the dataset is challenging even for state-of-the-art LLMs, which performed worse than the best student contestants, which had EM accuracy of 78.21\% and 377 points in the real-time proxy evaluation. The LLMs struggled most with Chemistry, as evidenced by the best EM equal to or less than 75\% compared with above 90\% for the other subjects. These results indicate that NSMQ is a challenging benchmark for current state-of-the-art LLMs, pointing to the need to improve LLMs on this kind of scientific and mathematical reasoning. This work contributes a novel and challenging benchmark for scientific and mathematical reasoning from the Global South towards enabling a true global benchmarking of LLMs capabilities for education.

\begin{credits}
\subsubsection{\ackname} We are grateful to ETH for Development (ETH4D) for funding this work with an ETH4D Research to Action Grant.
\end{credits}


\bibliographystyle{splncs04}
\bibliography{refs}

\end{document}